\documentclass{article}

\usepackage{microtype}
\usepackage{graphicx}
\usepackage{subfigure}
\usepackage{booktabs} 

\usepackage{times}
\usepackage{epsfig}
\usepackage{amsmath}
\usepackage{amssymb}

\usepackage{multirow}
\usepackage{color}
\usepackage{enumerate}
\usepackage{etoolbox}
\usepackage{mathrsfs}

\usepackage{hyperref}

\newcommand{\mypar}[1]{{\bf #1.}}
\usepackage[accepted]{icml2019}

\icmltitlerunning{Submission and Formatting Instructions for ICML LRG Workshop 2019}

\begin{document}

\twocolumn[
\icmltitle{Neural Message Passing 
for Visual Relationship Detection}



\icmlsetsymbol{equal}{*}

\begin{icmlauthorlist}
\icmlauthor{Yue Hu}{sjtu}
\icmlauthor{Siheng Chen}{merl}
\icmlauthor{Xu Chen}{sjtu}
\icmlauthor{Ya Zhang}{sjtu}
\icmlauthor{Xiao Gu}{sjtu}
\end{icmlauthorlist}

\icmlaffiliation{sjtu}{Cooperative Medianet Innovation Center, Shanghai Jiao Tong University, Shanghai, China}
\icmlaffiliation{merl}{Mitsubishi Electric Research Laboratories, Cambridge, Massachusetts, USA}

\icmlcorrespondingauthor{Ya Zhang}{ya{\_}zhang@sjtu.edu.cn}

\icmlkeywords{Machine Learning, ICML}

\vskip 0.3in
]



\printAffiliationsAndNotice{}  

\begin{abstract}
Visual relationship detection aims to detect the interactions between objects in an image; however, this task suffers from combinatorial explosion due to the variety of objects and interactions. Since the interactions associated with the same object are dependent, we explore the dependency of interactions to reduce the search space. We explicitly model objects and interactions by an interaction graph and then propose a message-passing-style algorithm to propagate the contextual information. We thus call the proposed method neural message passing (NMP). We further integrate language priors and spatial cues to rule out unrealistic interactions and capture spatial interactions. Experimental results on two benchmark datasets demonstrate the superiority of our proposed method. Our code is available at~\url{https://github.com/PhyllisH/NMP}.
\end{abstract}

\vspace{-5mm}
\section{Introduction}
\label{Introduction}
Visual relationship detection serves as a middle-level understanding task that bridges the gap between low-level image recognition, such as classification and object detection ~\cite{simonyan2014very, ren2015faster}, and high-level image understanding tasks, such as image captioning~\cite{vinyals2015show}, visual question answering~\cite{antol2015vqa}. Visual relationship denotes the visually recognizable interaction between subject and object, which is defined as triplet (\textit{subject-predicate-object}).
\begin{figure}[thp]
\vskip 0.1in
\begin{center}
\centerline{\includegraphics[width=1.0\linewidth]{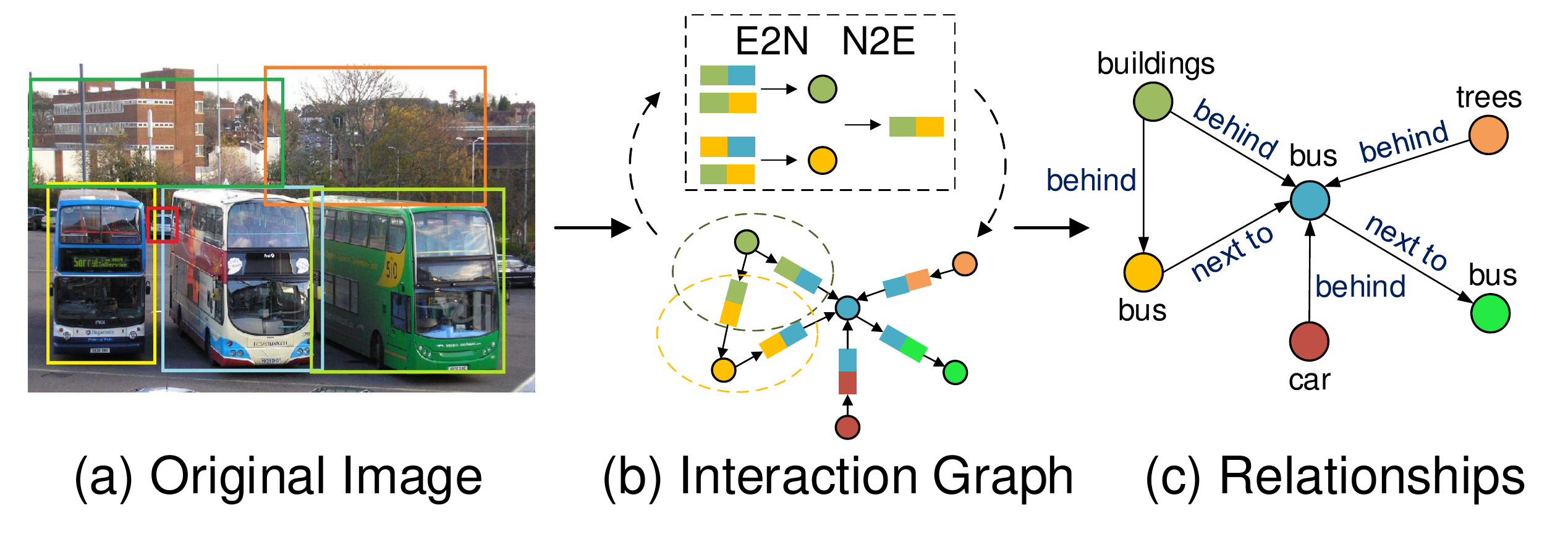}}
\caption{An interaction graph explicitly models objects and their interactions. We use message passing to propagate contextual information between objects and interactions to learn both node and edge embeddings. Pairwise relationships are detected based on edge embeddings.}
\label{fig: introduction}
\end{center}
\vspace{-10mm}
\end{figure}

\begin{figure*}[!htp]
\vspace{-2mm}
\begin{center}
\centerline{\includegraphics[width=0.95\textwidth]{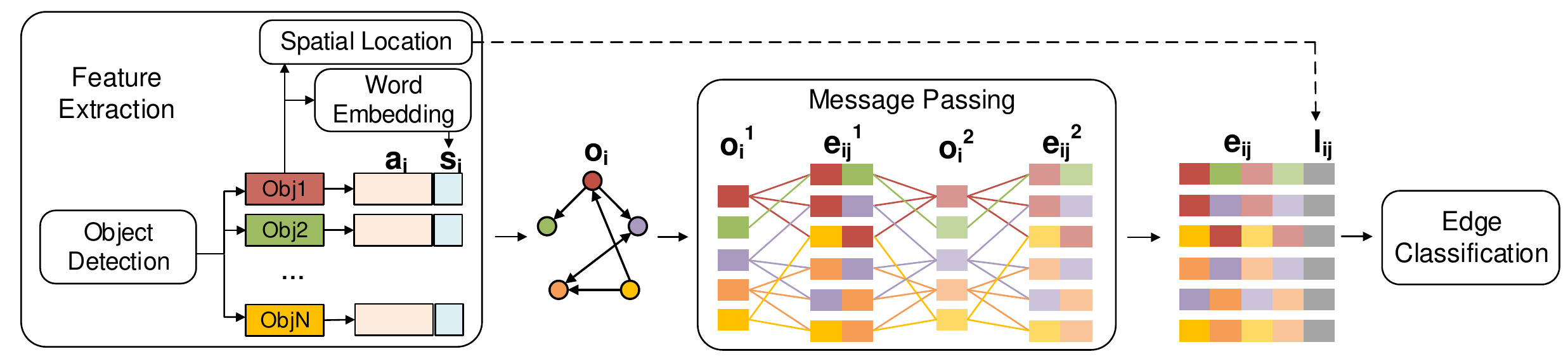}}
\caption{The overall framework of our proposed method, called neural message passing (NMP). Each detected object is represented by visual appearance and word embedding. A directed graph is built over these proposals, whose nodes denote the objects, edges denote the corresponding interactions. Message passing module is then applied to integrate contextual information. The concatenation of the enhanced interaction embeddings and the relative spatial locations are used for edge classification. Details can be found in Section \ref{section: method}.}
\label{fig: framework}
\end{center}
\vspace{-8mm}
\end{figure*}

Assuming there are $N$ object categories and $K$ predicate categories, there will be $N^2K$ relationship categories. The initial sequential mechanism treats each relationship triplet as a unique class and cannot apply to large dataset due to the explosive increase of the search space.~\cite{lu2016vrd} proposed a separation mechanism inferring objects and predicate separately, which reduces the complexity to $\mathcal{O}(N+K)$; however, this method leads to the missing context between objects and predicate. To address this,~\cite{li2017vip, yin2018zoom} proposes message passing within the relationship triplet to jointly extract features. Furthermore,~\cite{cui2018context} emphasizes global contexts by introducing the visual appearance of the surroundings; however, all the previous works ignore the relationship dependencies across relation triplets, \textit{e}.\textit{g}., the interaction between `bus' and `road' is more likely to be `park on' than `drive on' given `bus in the front of car' and `car park on road'. 

To exploit this contextual information, we construct an interaction graph for each image whose nodes denote the objects and edges denote the interactions. Different from the visual graph in~\cite{cui2018context}, which considers edges as an intermediate step to improve object embeddings, we explicitly use the edge embeddings to represent the relationship between objects. The edge embeddings are obtained through message passing over the interaction graph, which takes the high-order relationships into account. 
Considering visual appearance only may be difficult to capture all varieties of interactions. Semantic priors of objects and spatial locations are further introduced to rule out unreasonable interactions and capture spatial interactions.
The main contributions of this paper are: \\
$\bullet$  We propose a novel graph-based method to explicitly model interactions between objects in an image and use a message-passing-style algorithm to capture high-order interactions; \\
$\bullet$  We introduce the word embedding of each object and the relative spatial location between pairwise objects as the complement to the visual appearance; \\
$\bullet$  The proposed method consistently outperforms the previous state-of-the-art methods on two widely used datasets.

\section{Related Work}
Visual relationship detection has been extensively studied in recent years. At the very beginning, ~\cite{galleguillos2008object,farhadi2010every,sadeghi2011recognition,ramanathan2015learning} assigned a unique class to each relationship triplet; however, with the increase of objects and predicates, the amount of relationship triplets is explosive. To reduce the complexity,~\cite{lu2016vrd} learned objects and predicates separately; however, the separate model results in the lack of context between the related components. To address this,~\cite{yin2018zoom} encouraged feature sharing by message passing between the three components. Furthermore, the global contexts are introduced by utilizing graph.~\cite{liang2017deep} sequentially predicted interactions based on the semantic-action graph of the entire training set.~\cite{cui2018context} enhanced object embeddings by aggregating the visual appearance of the surroundings in the visual graph. However, the interaction embedding was ignored in the previous works. Instead, we explicitly model both interaction and object embeddings in the interaction graph. Language priors and spatial cues are further introduced to improve the performance in~\cite{plummer2017phrase, liang2018visual}. In this work, we integrate the word embeddings and spatial location to help estimate relationship.

Graph neural networks recently have got a lot of attention and achieved significant success in various fields~\cite{wang2017relational, gilmer2017neural, li2017situation, Battaglia2018RelationalIB, yang2018graph, niu2018generalized, woo2018linknet, kipf2018neural,abs-1810-05749}, especially in social networks~\cite{hamilton2017inductive}, knowledge graphs~\cite{kampffmeyer2018rethinking} and human object interaction~\cite{qi2018learning}. In this work, we apply graph neural networks to the application of visual relationship detection.

\section{Methodology}
\label{section: method}


\mypar{Overview}
Visual relationship detection has two settings: \textbf{predicate detection} and \textbf{relationship detection}. Predicate detection aims to predict the interactions between given pairs of objects. Relationship detection aims to simultaneously detect a set of objects and predict the interactions between pairs of objects.

We use multi-cues to better represent the objects and construct a graph to learn global contexts. The interaction graph organizes objects and interactions to structured data, such that we can jointly learn object and interaction embeddings. The main challenge is to explore the high-order interactions over structured data. To achieve this, we use a message-passing mechanism similar to~\cite{kipf2018neural}. The intuition is that each object is influenced by the related interactions and each interaction depends on the connected objects. The overview of the framework is in Fig. \ref{fig: framework}.

\mypar{Feature Extraction}
The functionality of this module is to get the visual and semantic cues of the objects and the relative spatial locations between pairwise objects. When only using visual appearance, the estimation of the predicate may be difficult due to the variety of relationships. While only using language priors, the relationship prediction is vague and not specified on the state of subject and object. Both visual and word embeddings represent each individual object, we further introduce relative spatial location to capture spatial interactions, such as 'near', 'under', 'on'.

The $i$-th object in the image is associated with a bounding box $b_i=\left\{x_i;y_i;w_i;s_i\right\}$ and a category $c_i$, which are given in the predicate detection task and obtained through object detection module in the relationship detection task. To extract deep visual features of an object, we adopt VGG16~\cite{simonyan2014very}. Firstly, we feed the original image into the network. When it comes to the last convolutional layer, we apply RoI align to crop out the bounding box, which is fed into the last fully connected layers afterward. The resulting features form the visual embedding $\mathbf{a_i}$. 
To complement the visual information, we use the pre-trained word2vector~\cite{mikolov2013efficient} to map the object category $c_i$ into word embedding $\mathbf{s_i}$. The object embedding of the $i$-th object $\mathbf{o}_i = [ \mathbf{a}_i;  \mathbf{s}_i ]$ is the concatenation of the visual embedding $\mathbf{a}_i$ and the word embedding $\mathbf{s}_i$. As for the spatial information, we adopt the idea of box regression and use box delta to get the box differences. Furthermore, we use intersection over union (iou) and normalized distance between two objects. The union bounding box of $b_i$ and $b_j$ is denoted as $b_{ij}$. $\mathbf{\Delta}(b_i, b_j)$~\cite{zhang2017relationship} is the box delta that regresses the bounding box $b_i$ to $b_j$. dis$(b_i,b_j)$~\cite{cui2018context} and iou$(b_i,b_j)$ denote the normalized distance and iou between $b_i$ and $b_j$. The spatial location between subject $v_i$ and object $v_j$ is $\mathbf{l_{ij}}=[\mathbf{\Delta}(b_i, b_j); \mathbf{\Delta}(b_i, b_{ij}); \mathbf{\Delta}(b_j, b_{ij}); \text{iou}(b_i, b_j); \text{dis}(b_i, b_j)]$.

\mypar{Graph Construction}
The interaction graph contains a node set $\mathbf{V}$ and an edge set $\mathbf{E}$.
Each node $v_i\in\mathbf{V}$ represents an object, which is composed of a bounding box $b_i$, a corresponding object category $c_i$ and the original embedding $\mathbf{o_i}$. Each edge $e_{ij}\in\mathbf{E}$ denotes the predicate between node $v_i$ and $v_j$. The relationship triplet ($v_i$-$e_{ij}$-$v_j$) and ($v_j$-$e_{ji}$-$v_i$) represent two different instances. To distinguish $e_{ij}$ from $e_{ji}$, we construct a directed graph.

For predicate detection, we construct a graph for each image based on the given object pairs. The edge exists when the connected objects are paired. We have observed that most of the interactions happen between close objects. For relationship detection, we assume that object interacts with the surroundings and assign the existence score of the edge based on dis$(b_i,b_j)$ and iou$(b_i,b_j)$ through the following formula, $t_1$ and $t_2$ are two thresholds.
\vspace{-1mm}
\begin{equation}
\begin{small}
{\rm exist}(e_{ij})=
\begin{cases}
1,& \text{dis(${b_i}$,${b_j}$) $<t_1$ or iou(${b_i}$,${b_j}$) $>t_2$},\\
0,& \text{otherwise}.
\end{cases}
\end{small}
\end{equation}
\vspace{-6mm}


\mypar{Neural Message Passing}
The functionality of the message passing module is to improve interaction embeddings by aggregating global context cues.
Instead of utilizing the common graph convolutional networks~\cite{kipf2016semi} to strengthen the node embeddings, we leverage node-to-edge and edge-to-node message passing mechanism similar to~\cite{gilmer2017neural, kipf2018neural} to explicitly model the node and edge embeddings. In the node-to-edge phase, each edge receives messages from the connected nodes. In the edge-to-node phase, the node embedding is updated according to the linked edge embeddings. Mathematically, the overall message passing module works as
\begin{align}
\label{original feature embedding} \mathbf{o_i^1}&=f_{\text{emb}}\left(\mathbf{o_i}\right) \\
\label{node2edge_1} v\rightarrow e: \mathbf{e_{ij}^1}&=f_e^1\left([\mathbf{o_i^1}; \mathbf{o_j^1}]\right) \\
\label{edge2node_1} e\rightarrow v: \mathbf{o_i^{2}}&=f_v^1\left([\frac{1}{d_i^{in}}\sum\limits_{e_{ji}\in \mathbf{E}}\mathbf{e_{ji}^1} ; \frac{1}{d_i^{out}}\sum\limits_{e_{ij}\in \mathbf{E}}\mathbf{e_{ij}^1}]\right) \\
\label{node2edge_2} v\rightarrow e: \mathbf{e_{ij}^{2}}&=f_e^{2}\left([\mathbf{o_i^{2}}; \mathbf{o_j^{2}}]\right) \\
\label{final edge fusion} \mathbf{e_{ij}}&=f_{\text{fusion}}\left([\mathbf{e_{ij}^1}; \mathbf{e_{ij}^2}]\right)
\end{align}
where $\mathbf{o_i}$ is the object embedding of the $i$-th object and $\mathbf{e_{ij}}$ is the edge embedding of between the $i$-th and $j$-th objects. The function $f_{\text{emb}}$ maps the original node embedding into the hidden space. Then we use the object embedding $\mathbf{o_i^1}$ to obtain edge embedding $\mathbf{e_{ij}^1}$. $[;]$ denotes concatenation. 
We use the concatenation rather than the sum or mean in order to distinguish the direction of the edges. $d_i^{in}$ is the amount of edges pointing to $v_i$, while $d_i^{out}$ is the amount of edges $v_i$ pointing out; both of which are used to normalize the edge embeddings. 
$\mathbf{e_{ij}^1}$ only depends on two node embeddings $\mathbf{o_i}$ and $\mathbf{o_j}$, while $\mathbf{e_{ij}^2}$ leverages more global information. Afterward, the final edge embedding $\mathbf{e_{ij}}$ in Eq.~\eqref{final edge fusion} is a fusion of the local embedding $\mathbf{e_{ij}^1}$ and the global embedding $\mathbf{e_{ij}^2}$. 

The functions ${f_e^1}$, ${f_v^1}$, ${f_e^2}$ are neural networks used for mapping between node and edge embeddings. In our experiments, we adopt two-layers fully-connected networks (MLPs) with Elu activation function which introduces non-linearity to enhance feature expression. 

\mypar{Edge Classification}
The functionality of this module is to classify the interactions between objects. The interaction embedding is the concatenation of the final edge embedding and spatial location; that is, $\mathbf{x_{i,j}} = [\mathbf{e_{ij}}; \mathbf{l_{ij}}]$. Then, the confidence of the predicate category between the $i$-th and the $j$-th objects is $y_{ij}= {\rm softmax} \left(\mathbf{W} \mathbf{x_{i,j}} \right),$
where $\mathbf{W}$ is the embedding matrix that maps interaction embeddings to match predicate categories. In our experiment, we use multi-class cross entropy loss for classification.


\begin{table}[h]
\vspace{-5mm}
	\caption{Predicate and relationship detection results (\%) in VRD dataset. "-" denotes the results are not reported in the original paper. $k$ denotes the number of predicates associated with each object. The total predicate category of VRD dataset is 70.}
	\label{tab:vrd-cmp}
	\vskip 0.1in
	\begin{center}
		\scriptsize
			\begin{sc}
				\begin{tabular}{cccccc}
					\toprule	
					\multirow{2}{*}{$k$} & \multirow{2}{*}{Methods} &\multicolumn{2}{c}{Predicate Det.} &\multicolumn{2}{c}{Relationship Det.} \\
					&&R@50 &R@100 &R@50 &R@100 \\
					\midrule
					\multirow{8}{*}{$k=1$} 
					&LP                 &47.87 &47.87 &13.86 &14.70      \\
					&VTE              &44.76 &44.76  &14.07  &15.20    \\
					&STA              &48.03 &48.03 &- &-               \\
					&CAI              &\underline{53.79}  &\underline{53.79}  &15.63 &17.39 \\
					&ViP-CNN        &-        &-        &17.32 &20.01    \\
					&Zoom-Net     & 50.69 &50.69  &\underline{18.92} &\underline{21.41} \\
					&VRL              &-        &-        &18.19  &20.79    \\
					&NMP            &\textbf{57.69} &\textbf{57.69} &\textbf{20.19} &\textbf{23.98} \\
					\midrule
					\multirow{5}{*}{$k=70$} 
					&DR-Net	      &80.78  &81.90  &17.73  &20.88    \\
					&Zoom-Net      &84.25  &90.59  & 21.37  &\underline{27.30} \\
					&DSR             &86.01 &93.18 &19.03 &23.29 \\
					&CDDN           &\underline{87.57} &\underline{93.76} &\underline{21.46} &26.14 \\
					&NMP            &\textbf{90.61} &\textbf{96.61} &\textbf{21.50} &\textbf{27.50}\\
					\bottomrule
				\end{tabular}
			\end{sc}
	\end{center}
	\vskip -0.2in
\end{table}

\begin{table*}[ht]
	\caption{Ablation Study (\%) on VRD dataset. \textbf{A} denotes the visual appearance of the object bounding box. \textbf{S} denotes the word embedding of the object category. \textbf{L} denotes the spatial location. \textbf{GRAPH} denotes contextual information through message passing.}
	\label{tab:abl}
	\vskip 0.1in
	\begin{center}
	    \scriptsize
		\begin{small}
			\begin{sc}
				\begin{tabular}{cccc|cccc}
					\toprule 	
					\multirow{3}{*}{Feature} &\multicolumn{3}{c}{Predicate Det.} &\multicolumn{4}{c}{Relationship Det.} \\
					& $k=1$ &\multicolumn{2}{c}{$k=70$} &\multicolumn{2}{c}{$k=1$} &\multicolumn{2}{c}{$k=70$}  \\
					&R@50/100 &R@50 &R@100 &R@50 &R@100 &R@50 &R@100  \\
					\midrule
					A &  48.98  & 86.24 & 94.34 & 17.92 & 21.59 & 19.29 & 25.20 \\
					A+L &  50.93 & 87.13 & 94.89 & 18.54 & 22.03 & 19.89 & 25.37 \\
					A+L+S &  53.60 & 89.57 & 96.19 & 18.49 & 22.27 & 20.30 & 26.14 \\
					\midrule
					Graph+A &  52.88 & 87.12 & 95.03 & 19.65 & 23.19 & 20.75 & 26.43 \\
					Graph+A+L&  54.13 & 88.71 & 95.64 & 19.99 & 23.51 & 21.48 & 26.90 \\
					Graph+A+L+S&  \textbf{57.69} & \textbf{90.61} & \textbf{96.61} & \textbf{20.19} & \textbf{23.98} & \textbf{21.50} & \textbf{27.50} \\
					\bottomrule
				\end{tabular}
			\end{sc}
		\end{small}
	\end{center}
	\vskip -0.1in
\end{table*}

\begin{table}[h]
\vspace{-5mm}
	\caption{Zero-shot predicate detection results (\%) in VRD dataset. Those methods without reporting the results on zero-shot setting are excluded from comparison.}
	\label{tab:zero}
	\vskip 0.1in
	\begin{center}
		\begin{small}
			\begin{sc}
				\begin{tabular}{cccc}
					\toprule
					\multirow{2}{*}{$k$} & \multirow{2}{*}{Methods} &\multicolumn{2}{c}{Predicate Det.} \\
					&&R@50 &R@100 \\
					\midrule
					\multirow{2}{*}{$k=1$} 
					&LP      &8.45   &8.45 \\
					&NMP    &\textbf{27.50}  &\textbf{27.50} \\
					\midrule
					\multirow{3}{*}{$k=70$} 
					&DSR     &60.90 & 79.81 \\
					&CDDN   &67.66 & 84.00 \\
					&NMP    &\textbf{72.95} & \textbf{88.44} \\
					\bottomrule
				\end{tabular}
			\end{sc}
		\end{small}
	\end{center}
\vspace{-8mm}
\end{table}

\begin{table}[h]
	\caption{Predicate detection results (\%) in VG dataset. The total predicate category of VG dataset is 100.}
	\label{tab:vg}
	\vskip 0.1in
	\begin{center}
		\begin{small}
			\begin{sc}
				\begin{tabular}{cccc}
					\toprule	
					\multirow{2}{*}{$k$} & \multirow{2}{*}{Methods} &\multicolumn{2}{c}{Predicate Det.} \\
					&&R@50 &R@100 \\
					\midrule
					\multirow{3}{*}{$k=1$}
					&VTE 	&62.63    &62.87 \\
					&STA    &62.71    &62.94 \\
					&NMP	&\textbf{67.03}    &\textbf{67.29} \\
					\midrule
					\multirow{4}{*}{$k=100$}
					&DSR          &69.06 & 74.37 \\
					&CDDN        &70.42 & 74.92 \\
					&DR-Net      &88.26 & 91.26 \\
					&NMP         &\textbf{89.69}  &\textbf{95.54}\\
					\bottomrule
				\end{tabular}
			\end{sc}
		\end{small}
	\end{center}
	\vskip -0.3in
\end{table}

\vspace{-3mm}
\section{Experiments}
\vspace{-1mm}
\mypar{Datasets}
\label{Datasets}
Visual Relationship Detection (VRD)~\cite{lu2016vrd} contains 5,000 images with 100 object categories and 70 predicate categories. There are 1,877 relationship triplets only exist in the test set, which is used for zero-shot evaluation. Visual Gnome (VG)~\cite{krishna2017visualGnome,zhang2017visual} contains 99,658 images with 200 object categories and 100 predicates. 

\mypar{Evaluation Metrics}
\label{Evaluation Metrics}
Following~\cite{lu2016vrd}, we use Recall@50 (R@50) and Recall@100 (R@100) as the evaluation metrics. R@n computes the fraction of true positive predicted relationships over the total annotated relationships among the top n confident predictions. Let $k$ be the number of predicates associated with each object. Similarly to~\cite{yu2017visual}, we report R@n under various $k$ values. 

\mypar{Compared with State-of-the-art Methods}
We compare our proposed model \textbf{NMP} against several previous state-of-the-art methods in Table~\ref{tab:vrd-cmp}: \textbf{LP}\cite{lu2016vrd}, 
\textbf{VTE}\cite{zhang2017visual},
\textbf{STA}\cite{yang2018shuffle},
\textbf{CAI}\cite{zhuang2017towards}, \textbf{DR-Net}\cite{dai2017detecting}, \textbf{ViP-CNN}\cite{li2017vip}, \textbf{Zoom-Net}\cite{yin2018zoom}, \textbf{VRL}\cite{liang2017deep}, \textbf{DSR}\cite{liang2018visual}, \textbf{CDDN}\cite{cui2018context}. 
We can see that (i) \textbf{NMP} consistently outperforms state-of-the-art methods under all settings; (ii) \textbf{NMP} outperforms \textbf{CDDN} by about 3\% according to Recall@100 on predicate detection task, which shows the effectiveness of our message passing algorithm over the directed interaction graph; (iii) we improve the state-of-the-art to 96.61\% and 27.50\% on predicate and relationship detection tasks.

Table~\ref{tab:zero} shows the comparison of the zero-shot predicate detection task. We exclude the performance on zero-shot relationship detection task since it is very sensitive to the number of detected bounding boxes. We can see that the performance is improved by about 5\% and 4.5\% on R@50 and R@100 respectively, which proves the promising generalization ability of our algorithm. To further prove the ability of our algorithm, we conduct experiments on the larger dataset: VG. Similarly to the previous works, we report results on the predicate detection task in Table~\ref{tab:vg}. Our algorithm achieves considerably superior performance than the previous works.

\mypar{Ablation Study}
We introduce message passing, visual embedding, word embedding and spatial location in our proposed network. Table~\ref{tab:abl} shows the influence of each factor to the performance. We test on both predicate and relationship detection tasks on VRD dataset. 
We see that (i) the spatial location and the word embedding improve the performance by around 1\% and 3\% respectively on predicate detection. Because the spatial information is not easy to learn from the visual appearance, and language priors can help rule out some obviously unreasonable compositions;  (ii) message passing ($\textbf{GRAPH}$) improves the recall stably by around 3\%, 2\% on predicate and relationship detection, respectively. The gain in relationship detection is smaller than that of predicate detection; this may be caused by the wrongly detected objects and incomplete annotation.

\section{Conclusions}
In this paper, we address the lack of context between interactions in previous works. We construct an interaction graph with a message-passing mechanism to explore high-order interactions. Besides, we use visual appearance, language priors, and spatial cues to complement each other. Experimental results show that our proposed method outperforms the state-of-the-art methods on two benchmark datasets.


\bibliography{example_paper}
\bibliographystyle{icml2019}




\end{document}


\twocolumn[
\icmltitle{Neural Message Passing 
for Visual Relationship Detection}



\icmlsetsymbol{equal}{*}

\begin{icmlauthorlist}
\icmlauthor{Yue Hu}{equal,to}
\end{icmlauthorlist}

\icmlaffiliation{to}{School of Electronic Information and Electrical Engineering, Shanghai Jiao Tong University, Shanghai, China}

\icmlcorrespondingauthor{Yue Hu}{18671129361@sjtu.edu.cn}

\icmlkeywords{Machine Learning, ICML}

\vskip 0.3in
]



\printAffiliationsAndNotice{}  

\section{Implement Details}
We fine-tune Faster R-CNN on the bounding boxes in the train set of VRD and VG. Afterward, we feed the fine-tuned parameters into VGG16 to extract visual features. Then we fix the VGG16 parameters in the following steps. We adopt RMSProp optimizer to optimize the whole network and the learning rate is set to be 0.0005. In the relationship detection task, the distance threshold $t_1$ is set to be 0.45 and the iou threshold $t_2$ is 0.50.